\documentclass[lettersize,journal]{IEEEtran}
\usepackage{amsmath,amsfonts}
\usepackage{algorithmic}
\usepackage{algorithm}
\usepackage{array}
\usepackage[caption=false,font=normalsize,labelfont=sf,textfont=sf]{subfig}
\usepackage{textcomp}
\usepackage{stfloats}
\usepackage{url}
\usepackage{verbatim}
\usepackage{graphicx}
\usepackage{cite}
\begin{document}

\title{A Graph-Enhanced DeepONet Approach for Real-Time Estimating Hydrogen-Enriched Natural Gas Flow under Variable Operations}

\author{Sicheng Liu, Hongchang Huang, Bo Yang, Mingxuan Cai, Xu Yang, Xinping Guan,~\IEEEmembership{Fellow,~IEEE}
        % <-this % stops a space
\thanks{This work is supported by the National Natural Science Foundation of China (No. 62273237, 62325306).}}% <-this % stops a space



\maketitle

\begin{abstract}
Blending green hydrogen into natural gas presents a promising approach for renewable energy integration and fuel decarbonization. 
Accurate estimation of hydrogen fraction in hydrogen-enriched natural gas (HENG) pipeline networks is crucial for operational safety and efficiency, yet it remains challenging due to complex dynamics. 
While existing data-driven approaches adopt end-to-end architectures for HENG flow state estimation, 
their limited adaptability to varying operational conditions hinders practical applications.
To this end, this study proposes a graph-enhanced DeepONet framework for the real-time estimation of HENG flow, especially hydrogen fractions. 
First, a dual-network architecture, called branch network and trunk network, is employed to characterize operational conditions and sparse sensor measurements to estimate the HENG state at targeted locations and time points.
Second, a graph-enhance branch network is proposed to incorporate pipeline topology, improving the estimation accuracy in large-scale pipeline networks.
Experimental results demonstrate that the proposed method achieves superior estimation accuracy for HCNG flow under varying operational conditions compared to conventional approaches.
\end{abstract}

\begin{IEEEkeywords}
Hydrogen-enriched natural gas, hydrogen fraction, state estimation, DeepONet.
\end{IEEEkeywords}

\section{Introduction}
\IEEEPARstart{B}{lending} green hydrogen into natural gas pipelines offers a promising solution for integrating renewable energy and decarbonizing fossil fuels.
Moreover, the resulting hydrogen-enriched natural gas (HENG) pipeline system effectively mitigates costly infrastructure investments, particularly in city areas.

\begin{figure}[!t]
% \centerline{\includegraphics[width=\columnwidth]{fig_configure}}
\centerline{\includegraphics[width=\columnwidth]{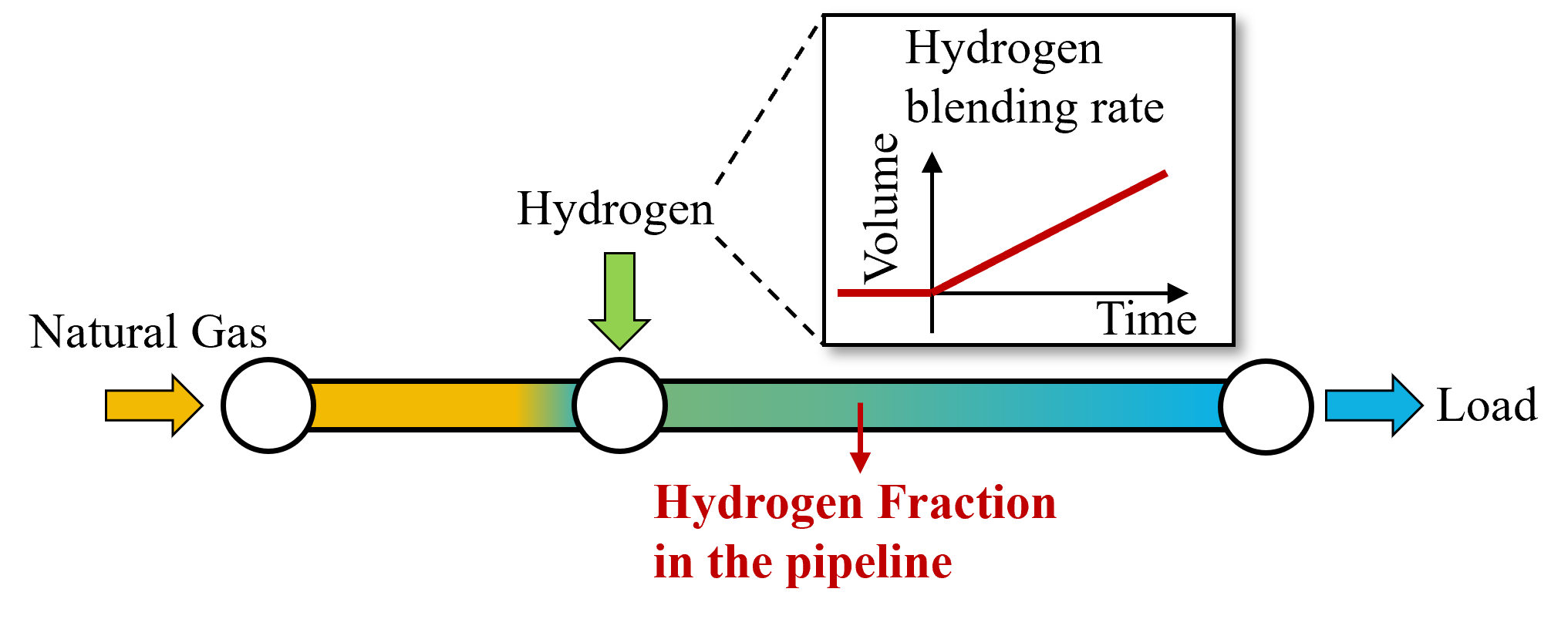}}
\caption{Illustration for hydrogen fractions of HENG in the pipeline network.}
\label{fig_back}
\end{figure}

A critical challenge in HENG pipeline network operation involves accurate flow estimation, especially hydrogen fraction estimation throughout the pipeline network.
As illustrated in Fig. \ref{fig_back}, adjustments of operations such as hydrogen blending flow rate and nodal pressure, dynamically alter hydrogen distributions, resulting in non-uniform concentration profiles along pipeline networks.
This spatial-temporal variability directly impacts system performance, since the hydrogen concentration not only determines the gas heating value.
(directly governing industrial energy consumption rates) but also introduces safety risks when exceeding permissible thresholds \cite{chen2020LinJin}.
% For instance, excessive fractions may trigger combustion instability in end-use equipment.
Therefore, precise hydrogen fraction estimation plays an important role in ensuring both economic efficiency and operational safety in HENG systems.
% Hydrogen fraction directly governs the gas heating value, determining energy demand in industrial applications. 
% Exceeding permissible thresholds risks combustion instability in transport and end-use devices.
% Hydrogen fractions not only influence the gas calorific value, directly affecting consumption volumes, but exceeding permissible limits may compromise equipment safety.

In fact, the hydrogen fraction dynamics in pipelines are governed by partial differential equations (PDEs) \cite{mhanna2022coordinated}.
% The highly nonlinear nature of PDEs forces current HENG scheduling studies, whether employing static \cite{isam2021JieDian,yunpeng2024ChongDa} or dynamic formulations \cite{2025HaoXianQingDong, sleiman2022YouHVF}, to adopt empirical simplifications or thermodynamic relaxations for computational feasibility.
Due to the highly nonlinear nature of PDEs, current HENG scheduling studies based on whether static \cite{isam2021JieDian,yunpeng2024ChongDa} or dynamic formulations \cite{2025HaoXianQingDong, sleiman2022YouHVF} adopt simplifications or relaxations for computational feasibility.
% employ simplifications or relaxations to maintain computational tractability.
This leads to a gap between scheduled and actual states about HENG flow, compromising control reference accuracy and safety margin evaluations.

% Since the complexity of hydrogen transport dynamics, current HENG operational models, including both static \cite{isam2021JieDian,yunpeng2024ChongDa} and dynamic formulations \cite{2025HaoXianQingDong, sleiman2022YouHVF}, often employ simplifications or relaxations to maintain computational tractability.
% This underscores the need for HENG flow estimation in post-decision, i.e., under variable operations that enable precise equipment control while ensuring hydrogen concentration safety margins.

% Due to the complexity of hydrogen dynamics, existing studies on HENG system operation scheduling have often relied on static \cite{} and dynamic \cite{} models that are simplified or relaxed to improve computational efficiency. 
% This simplification highlights the need for more refined post-decision hydrogen fraction estimation, which provides better references for equipment control and safety verification.

% In practice, the hydrogen fraction dynamics in pipelines follow a set of partial differential equations (PDEs) \cite{mhanna2022coordinated}. 
% Traditional solution methods, such as the finite element method or finite volume method, are computationally expensive. 
% Furthermore, due to sensor deployment limitations, some system states remain unobservable. 

% In fact, the hydrogen fraction dynamics in pipelines are governed by partial differential equations (PDEs) \cite{mhanna2022coordinated}.
This limitation necessitates precise HENG flow estimation in post-scheduling.
Facing the difficulty of computational inefficiency and requiring full boundary/initial conditions through traditional numerical PDE-solving methods, 
some studies explored physics-informed neural networks (PINNs) for real-time pipeline dynamics estimation, including HENG \cite{zhang2025physics}, liquid \cite{du2024deeppipe}, and gas flow \cite{su2018systematic}.
However, conventional PINNs are mainly suitable for fixed initial/boundary conditions of PDEs, limiting their applicability to the HENG system with variable operations.
% Conventional numerical solving methods for PDEs such as finite-volume methods face computational bottlenecks and depend on extensive sensor deployments.
% To this end, some studies explored physics-informed neural networks (PINNs) for pipeline dynamics estimation, including HENG \cite{zhang2025physics}, liquid \cite{du2024deeppipe}, and gas flow \cite{su2018systematic}.

% These challenges have motivated research into using end-to-end neural networks to solve PDEs and model pipeline dynamics. 
% For example, Zhang et al. proposed a physics-informed neural network (PINN) for estimating the state of hydrogen-containing natural gas pipeline systems \cite{zhang2025physics}, while customized PINN-based methods have been developed for liquid and natural gas pipelines \cite{du2024deeppipe,su2018systematic}.
% However, PINNs are typically designed for solving PDEs with fixed initial and boundary conditions, making them less adaptable to varying operational conditions.

Recently, the DeepONet framework offered an improvement in PDE approximation, allowing for varying initial/boundary conditions as inputs through branch networks in its architecture \cite{lu2021Learning}.
Thus DeepONet provides a promising method for HENG flow estimation under variable operations.
% enhanced operator learning capabilities, including PDEs, under varying conditions through its dual-network architecture: 
% Branch networks encode input functions (e.g., boundary conditions) while trunk networks capture spatial dependencies \cite{lu2021Learning}.
% Thus DeepONet provides a promising method for HENG flow estimation under variable operations.
Nevertheless, the estimation accuracy of classical DeepONet approaches may degrade for large pipeline networks, as they require separate branch networks for each condition with multiplicative combinations.
In the HENG system, we observe that the conditions of each pipeline are only affected by neighboring pipelines, thus there is no need for direct aggregation of the global branch net features directly.
This motivates the integration of graph neural networks into DeepONet's branch network architecture.
% Our key insight recognizes that pipeline network topology creates localized condition dependencies.
% To address this limitation, the DeepONet framework provides a promising alternative for hydrogen fraction estimation under dynamic conditions. 
% DeepONet employs two neural networks—the branch net and trunk net—to capture spatial characteristics of functional problems and input function features, respectively.
% Nevertheless, traditional DeepONet requires a separate branch net for each initial or boundary condition, leading to a rapid increase in model size and reduced learning efficiency as the HENG pipeline network grows. 
% Observing that only neighboring pipelines influence each other due to network topology, we propose integrating graph neural networks into DeepONet to improve estimation effectiveness.

In summary, this paper proposes a graph-enhanced DeepONet framework for hydrogen fraction estimation in HENG pipeline networks under variable operations. The detailed contributions are as follows:
\begin{enumerate}
    \item {
    Facing the challenges of variable operations for data-driven HCNG flow estimation, a DeepONet-based approach is established. 
    Different from existing PINN-based approaches like \cite{zhang2025physics}, the proposed approach introduces branch nets to incorporate partial initial and boundary conditions of the pipeline networks, enhancing its estimation performance under variable operations.
    % enables the incorporation of initial and boundary conditions, making it suitable for variable HENG system operations.
    }
    \item{
    A graph-enhanced structure is developed for DeepONet, which leverages the topology of the pipeline network to enhance the aggregation of branch nets.
    This structure achieves higher parameter efficiency compared with conventional classical DeepONet \cite{lu2021Learning}, thus enhancing the scalability of the approach on large-scale pipeline networks.
    % incorporating pipeline network topology for branch 
    % DeepONet architecture is developed, incorporating pipeline network topology through a graph-based branch network. 
    % This enhances the representational efficiency and estimation accuracy compared with conventional classical DeepONet \cite{lu2021Learning}, especially for the large-scale pipeline network.
    }
\end{enumerate}

\section{Brief Methodology}
\subsection{Dynamics of HENG Flow}
The dynamics of HENG flow can be described by a set of dynamic equations.
Due to space constraints, for example, the dynamic of hydrogen fractions can be simplified and expressed as
% \begin{spacing}{0.7}
\begin{equation}
\label{deqn_ex1}
{\frac{\partial w}{\partial t} + 
\frac{m}{A\rho}
%\frac{\overline{Q}_{ij}}{\overline{\rho}_{ij}}
\frac{\partial w}{\partial x} = 0}
\end{equation}
where variables $w$, $\rho$, and $m$ represent the hydrogen mass fraction, density, and mass flow rate of HENG in pipeline $ij$, respectively. Parameter $A$ is the cross-sectional area of pipeline $ij$.

Moreover, the boundary and initial conditions of Eq. (1) also affect the hydrogen fraction.  
Specifically, the boundary conditions $w_{ij}|_{x=0}$ of pipelines with specific components are determined by their operations, such as gas sources and hydrogen stations.
And partial initial conditions $w_{ij}|_{t=0}$ can be acquired by a limited number of sensors.
% [XXXXXX]%这导致了估计器对有限的边界条件的输入需求。
This information is critical for the accuracy of HENG flow estimation, thus leading to the introduction of the following DeepONet structure.

% Therefore, it is necessary to consider these conditions for HENG estimation under different operations.

\subsection{Proposed Graph-Enhanced DeepONet Structure}
\begin{figure}[!t]
% \centerline{\includegraphics[width=\columnwidth]{fig_configure}}
\centerline{\includegraphics[width=\columnwidth]{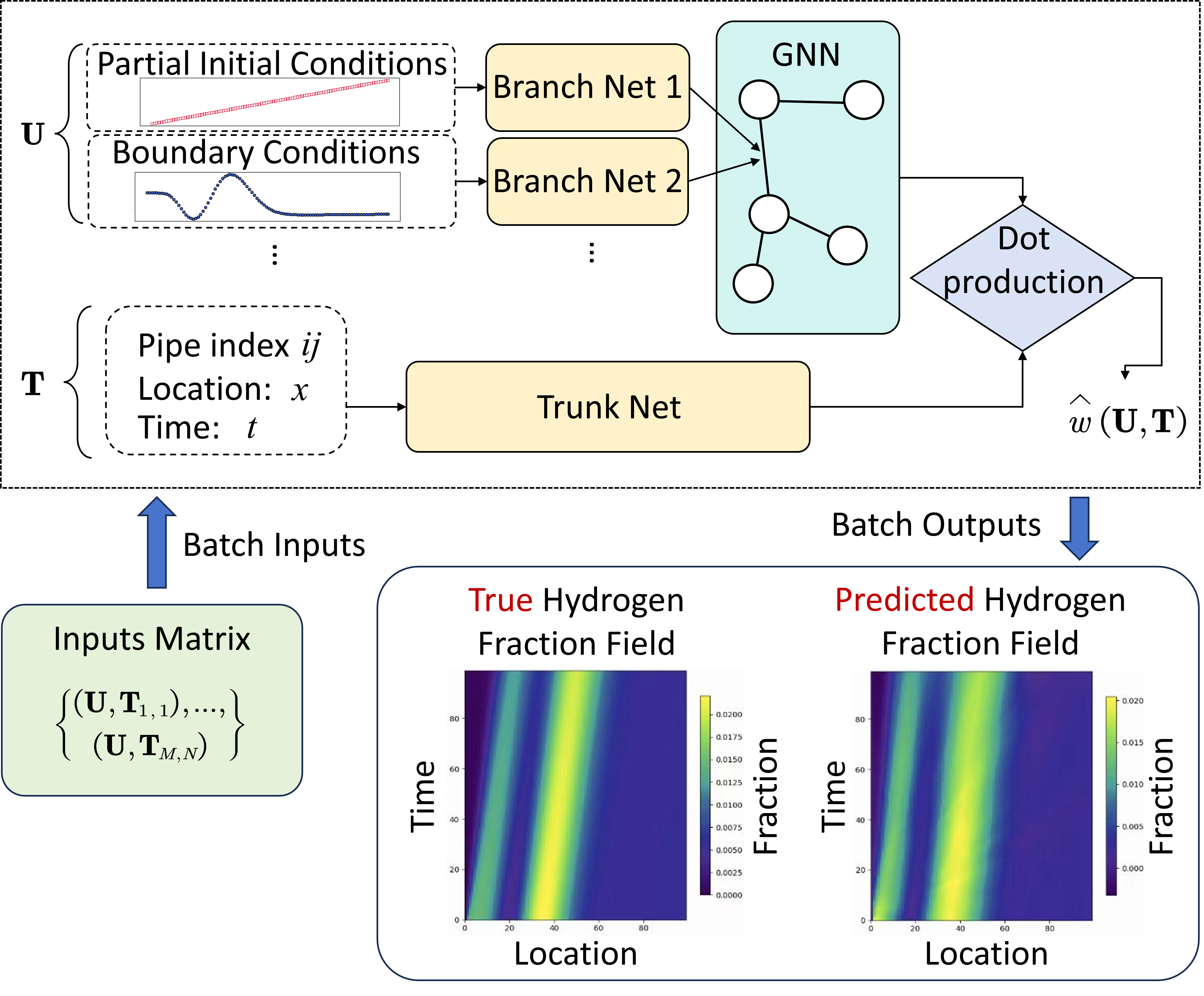}}
\caption{Framework of the proposed graph-enhanced DeepONet for HENG flow estimation.}
\label{fig_DeepONet}
\end{figure}

The structure of the proposed graph-enhanced DeepONet is shown in Fig. \ref{fig_DeepONet}.
Branch nets are used to characterize partially known boundary and initial conditions of HENG flow, and subsequently aggregates through a graph neural network.
% spatial characteristics and input function features, i.e., the boundary and initial conditions of pipelines. 
The trunk net is used to input the desired location and moment of HENG flow.
The outputs of the above two partials are subsequently combined through a dot-product operation to estimate the state of HENG flow such as hydrogen fraction.

In concrete, the graph-enhanced DeepONet is established to estimate: $\mathcal{G}: (\mathbf{U},\mathbf{T}) \to \widehat{w}$.
The inputs of branch nets are $\mathbf{U}=\{\boldsymbol{u}_{ij}^{\mathrm{init}},\boldsymbol{u}_{ij}^{\mathrm{bound}}\}$.
Vector $\boldsymbol{u}_{ij}^{\mathrm{init}} = [u_{ij}^{\mathrm{init},1},...,u_{ij}^{\mathrm{init}, S}]$ represents the partial initial condition acquire by the limited sensors, where $S$ is the number of sensors.
Vector $\boldsymbol{u}_{ij}^{\mathrm{bound}} = [u_{ij}^{\mathrm{bound},1},...,u_{ij}^{\mathrm{bound}, K}]$ represents the boundary condition determined by the operations, where $K$ is the number of samples.
The input of the trunk net $\mathbf{T}=\{ij,x,t\}$, where $ij$ is the index of pipelines, $x$ is the relative position of pipeline $ij$, and $t$ is the moment.
The output of the graph-enhanced DeepONet is the estimated HENG state $\widehat{w}$, such as hydrogen fraction, at $\{ij,x,t\}$ under condition $\mathbf{U}$. 
The mean square errors from estimated states $\widehat{w}$ with the real values $w$ are used as loss for training the DeepONet.

Moreover, the graph neural network is integrated.
Since the topology of the HENG network, the conditions of each pipeline only affect neighboring pipelines.
The topology-aware structure enhances the representational ability and parameter efficiency of DeepONet for estimating the HENG flow, especially for the large-scale pipeline network in which the pipeline and branch net number are large.

\bibliographystyle{IEEEtran}
\bibliography{Paper3}

\vfill

\end{document}